\documentclass[runningheads]{llncs}
\usepackage{graphicx}
\usepackage{amsmath,amssymb} % define this before the line numbering.
\usepackage{color}

\begin{document}
% \pagestyle{headings}
% \mainmatter

% \def\ACCV18SubNumber{778}  % Insert your submission number here

%===========================================================
% \title{FineTag: Multi-label Retrieval of Attributes at Fine-grained Level in Images} % Replace with your title
\title{FineTag: Multi-attribute Classification at Fine-grained Level in Images} % Replace with your title
% \titlerunning{ACCV-18 submission ID \ACCV18SubNumber}
% \authorrunning{ACCV-18 submission ID \ACCV18SubNumber}
\titlerunning{FineTag}

\author{Roshanak Zakizadeh, Michele Sasdelli, Yu Qian and Eduard Vazquez}
\authorrunning{R. Zakizadeh et al.}

\institute{Cortexica Vision Systems, London, UK\\
\email{roshanak.zakizadeh@cortexica.com}}

\maketitle

%===========================================================
\begin{abstract}
% The abstract should summarize the contents of the paper and should
% contain at least 70 and at most 300 words. It should be set in 9-point
% font size and should be inset 1.0~cm from the right and left margins.

%In image retrieval, the features extracted from an item are used to look for  similar lookalike of the items (e.g. finding a match for a bag in a retail catalogue). Identifying all the attributes of a single instance of an item in an image (e.g. the shape of the bag) could facilitate the task of image retrieval. 
In this paper, we address the extraction of the fine-grained attributes of an instance as a ``multi-attribute classification'' problem. 
To this end, we propose an end-to-end architecture by adopting the bi-linear Convolutional Neural Network with the pairwise ranking loss. This is the first time such architecture is applied for the fine-grained attributes classification problem. We compared the proposed method with a competitive deep Convolutional Neural Network baseline. Extensive experiments show that the proposed method attains/outperforms the performance of compared baseline with significantly less number of parameters ($40\times$ less). We demonstrated our approach on CUB200 birds dataset whose annotations are adapted in this work for multi-attribute classification at fine-grained level.

%Fine-grained categorization where an item is classified within a minor category (e.g. recognizing species of dogs) is the closest area of research to multi-label attributes classification of an \textit{instance}. The recently proposed bi-linear convolutional neural network is shown to be very efficient in fine-grained categorization. Inspired by this architecture, here we propose a solution (FineTag) for  multi-label attributes recognition at the fine-grained level. In our design, we applied the state-of-the-art pairwise ranking based loss function for multi-label classification. Further, we adopt the challenging CUB200 birds dataset for the task of multi-label attributes classification at fine-grained level\textcolor{red}{(is it done for the first time? Have you annotated by yourself? Then it should be highlighted as one of the contribution. A small section experiment describing this data-set increases the chance of acceptance)}. We compare our results with a competitive deep CNN architecture (vgg16) and show that it is possible to achieve the comparable or even better performance with a significantly less number of parameters ($40\times$ less).

% \dots
\end{abstract}

%===========================================================
\section{Introduction}

\begin{figure}
	\centering
	\includegraphics[trim={5cm 12cm 3cm 2cm},clip, width=\linewidth]{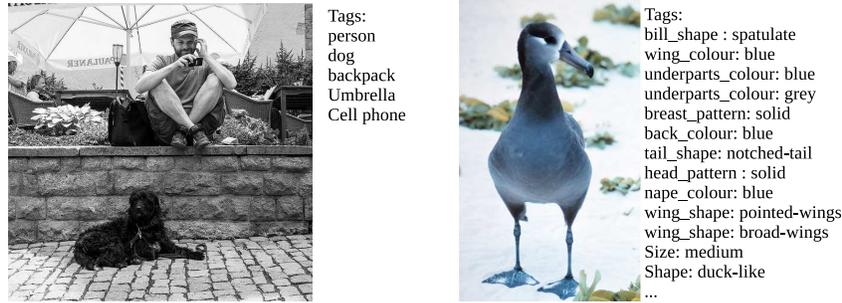}
	\caption{On the left: multi-label image classification:tagging an image at high level of recognition (the image is from COCO datasets~\cite{lin2014microsoft}). On the right: multi-attribute \textbf{instance} classification:tagging an object at fine-grained level of recognition (the image is from CUB200 bird dataset~\cite{wah2011caltech}).}
	\label{fig:teasor}
\end{figure}

%The most common computer vision problem which is addressed with regards to the fine-grained data is fine-grained categorization. 
Fine-Grained Visual Categorization (FGVC) is an important and an active research topic in computer vision~\cite{lin2015bilinear,cui2018large}. It is the task of identifying objects from subordinate categories, for example, recognizing species of birds~\cite{lin2015bilinear}. Unlike fine-grained categorization, in attributes recognition at fine-grained level we are interested in classifying the attributes of an instance. %\textcolor{red}{Add one sentence mentioning the usefulness of attributes recognition at fine-grained level eg, such classification is quite useful in fine grained product search...} 
An example of this can be seen in Figure~\ref{fig:teasor} where an instance of a specific bird is classified for different attributes such as the bill shape or the under-part color. Retrieving such attributes of objects have been shown to be effective in improving object recognition and categorization \cite{berg2010automatic}. Further, given the fact that such higher-order description of the images can provide semantically meaningful information ~\cite{berg2010automatic,farhadi2009describing}, along with low-level features (whether hand-crafted or extracted from different layers of a CNN) they can facilitate the task of fine-grained retrieval for instance for product search~\cite{songfine,wang2014learning}.  Specifically, it is been noted~\cite{wang2014learning} that it is worthwhile to learn a fine-grained model that is capable of characterizing the fine-grained visual similarity for the images within the same category.

% One major application of the recognized attributes of a fine-grained item is in retrieval where the they are used to retrieve the exact individual item of query from a dataset~\cite{rui1999image,wang2014learning}. 

Previous work have addressed the problem of recognizing attributes of an object at fine-grained level using other techniques such as relying on text information to associate attributes with the object, for instance describing jewelries or other fashion items with fine-grained attributes such as color or texture~\cite{berg2010automatic}. Some work have taken advantage of localizing the object or the fine parts of the object~\cite{duan2012discovering,zhang2013deformable,berg2013poof,wei2017selective} for example in~\cite{duan2012discovering} the discriminative visual attributes are detected for birds images. %\textcolor{red}{Most of these work have targeted fine-grained categorization as the final application of extracted attributes and the reported results are not showing the recognition precision per attribute,... this point is not clear. Is it so difficult to get per attribute performance from their reported results?}. 
In this work, we demonstrate that retrieving attributes of an object at a fine-grained level can be looked at as a multi-attribute \textit{attribute} classification task. As opposed to the multi-attribute \textit{object} classification (the example on the left in Figure~\ref{fig:teasor} taken from COCO dataset~\cite{lin2014microsoft} where the scene is classified for multiple objects at high level of recognition), in \textit{multi-attribute} classification a single instance of an object will be tagged for its different attributes at a fine-grained level of recognition (see the bird example on the right in Figure~\ref{fig:teasor} which is tagged for the shape of its bill, its under-part colors, etc.). However, we can take advantage of the methods in multi-attribute image classification domain. Pairwise ranking loss has been proven to be successful in multi-label image classification with annotation of tags as the ground truth \cite{weston2011wsabie,gong2013deep}. The ranking objective in the pairwise ranking loss function is that the score of the positive labels is larger than the score of negative labels. The derivative of the ranking loss does not converge to zero when the logits have high values. This makes it robust against the vanishing gradient problem. Moreover, similar to multi-label object classification, in multi-attribute classification we are interested in the per attribute accuracy results.

As we mentioned before, fine-grained categorization is an area of research which is very close to attributes extraction at the fine-grained level. The Fine-grained categorization problem has been addressed using different techniques including part-based or localization models ~\cite{zhang2014part}. However, these techniques often require additional expensive annotations such as bounding boxes \cite{zhang2014part} or keypoints \cite{liu2016deepfashion}. Among the recent techniques bilinear convolutional neural networks (bcnn)~\cite{lin2015bilinear} have shown to improve the results for fine-grained categorization significantly. In bcnn~\cite{lin2015bilinear} an image is passed through two (similar or different) convolutional networks and the outputs of their last convolutional layer (more accurately conv+pool) are multiplied using the outer product at each location of the image and sum-pooled to obtain the bilinear vector. This additional step is defined as the bilinear-pool layer. In our proposed FineTag architecture we have adopted the idea of bilinear-pool layer to allow capturing a feature map with an emphasis on fine details and use it not as a second step of training but as convolutional layer in the network to enable end-to-end training.

Our contributions in this paper are as follows: 1. We introduce FineTag architecture which is a simple network (in terms of the number of parameters)  to retrieve the attributes of an instance at fine-grained level of details. 2. We adapt CUB200 birds dataset by Caltech~\cite{wah2011caltech}, which was initially collected and organized for fine-grained categorization, to be used for multi-attribute classification of attributes at fine-grained level. 

The paper is organized as follows: In Section~\ref{sec:method}, we describe the architecture of our network (FineTag) including the loss function used by the model as well as the evaluation metrics used. Section~\ref{sec:data} discusses preparing the data for the experiment, as well as the challenge of data in Fine Tagging. In Section~\ref{sec:exp}, the experiments and results are discussed and Section~\ref{sec:sum} summarizes our work and outlines the future challenges and goals.

% repeat the examples in the abstract here

% This document serves as an example submission. It illustrates the format
% we expect authors to follow when submitting a paper to ACCV. 
% At the same time, it gives details on various aspects of paper submission,
% including preservation of anonymity and how to deal with dual submissions,
% so we advise authors to read this document carefully.
%Do not use any additional Latex macros.

%------------------------------------------------------------------------- 
\section{Method}
\label{sec:method}
\subsection{FineTag Network Architecture}
\begin{figure}[!h]
    \centering
    %trim={<left> <lower> <right> <upper>}
    \includegraphics[trim={0cm 8cm 0cm 0cm},clip, width=\linewidth]{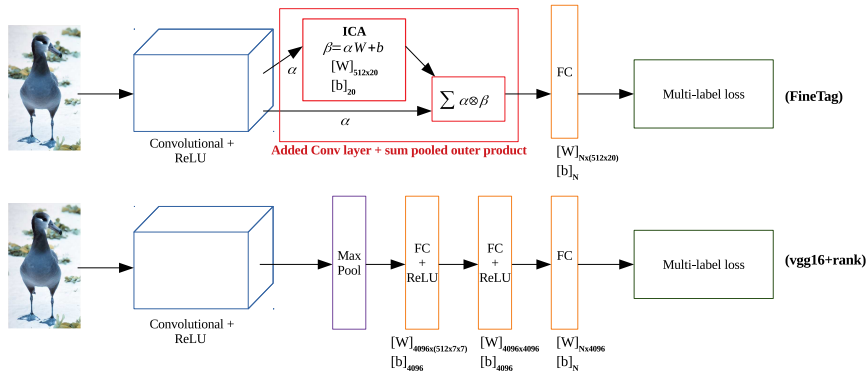}
    \caption{\textbf{FineTag}: a Multi-label retrieval network for extracting the attributes at fine-grained level in an image (the top network), in comparison with VGG16+rank \cite{simonyan2014very} architecture (in the bottom). The number of parameters in the proposed architecture is significantly less than VGG16+rank (i.e. $40\times$ less).}
    \label{fig:net}
\end{figure}

Figure~\ref{fig:net} shows the architecture of our proposed model (the graph in the first row). Our proposed architecture is fully convolutional \cite{long2015fully} and it is based on VGG16 \cite{simonyan2014very}. VGG16 is chosen due its small filter kernel size ($3\times3$) compared to other existing convolutional neural networks which makes it suitable for capturing fine texture details in an object. We applied the bilinear-pool layer in~\cite{lin2015bilinear} as an outer product layer and used a loss function which is suitable for multi-label classification proposing an end-to-end model for multi-attribute extraction of an instance at a fine-grained level.

Similar to~\cite{lin2015bilinear}, we use VGG16 pre-trained on the ImageNet dataset \cite{deng2009imagenet} truncated at a convolutional layer after the non-linearities. More specifically, we extract the features from the last convolutional layer of VGG16 with non-linearities, i.e. layer 30 ($conv_{5\_3}$ + $relu$).

The size of the feature map ($\alpha$) at this level is  $512\times14\times14$. We project a copy of $\alpha$ into a 20 dimensional ICA \cite{hyvarinen1999survey} projection space which is generated from the feature maps extracted from the same $30^{th}$ layer when a set of images from the same dataset are passed through the VGG16 network. This results into a reduced feature map ($\beta$) of size $20\times14\times14$.  We also tried PCA for dimensionality reduction, however we found ICA to be performing at least $1\%$ better than PCA (at  least with CUB200 bird dataset~\cite{wah2011caltech}).

Then, the sum of the outer product of $\alpha$ and $\beta$ at each location is calculated which results in a $512\times20\times14\times14$ feature map. Summing over the spatial dimensions this is reduced to $512\times20$ features. As it can be seen in Figure~\ref{fig:net}, the ICA projection can be designed as a $1\times1$ convolutional operation with weights and biases of size $[512, 20]$ and $[20, 1]$ respectively. This architecture relies on the local correlation between pairs of features for creating the final list of logits necessary for multi-labelling.
The layer following the sum of the outer product is a fully connected layer without any non-linearity (similar to the last fully connected layer in the original VGG16 design). To get the predicted tags for the image the results are passed through a  multi-label classifier with the multi-label ranking loss during training (shown in green in Figure \ref{fig:net}).

The number of classes (denoted as $N$ in Figure~\ref{fig:net}) is the size of the vocabulary of attributes. For instance, in case of CUB200 bird dataset \cite{wah2011caltech} where there are 312 attributes in total, the network is trained for an  output of 312 logits. The network is trained with a multi-labelling ranking loss (Eq.~\ref{eq:smooth_rank_loss}). The rank of the logits gives the tagging results. An optimally trained network would rank the labels present in the image higher than the ones missing.

An important advantage of our proposed architecture is that it requires much less parameters to train compared to the baseline network: very deep VGG16. Figure~\ref{fig:net} provides a visual comparison of the weights and biases in the layers of our proposed network and of VGG16+rank. Apart from the number of parameters shared between the two models in the convolutional $+$ ReLU layers (roughly  $1.6\times10^{6}$), with $N=312$ (number of attributes in CUB200 bird dataset) our model required about $3\times10^{6}$ parameters to train, whereas, VGG16 requires around $120\times10^{6}$ parameters to train for the same number of attributes (i.e. FineTag has $40\times$ less parameters). As it is predictable by the number of parameters and the extra fully connected layers in VGG16, FineTag is more than 30\% faster to train. 

Another advantage of our proposed architecture is that, unlike what is usual with fine-grained recognition techniques~\cite{duan2012discovering,berg2013poof}, with FineTag we do not have to include explicitly the location of the parts in the object.

We emphasis FineTag is and end-to-end architecture and the ICA projection space coefficients which are only calculated once beforehand and used for initialization of bilinear layer (shown in red in Figure~\ref{fig:net}).

\subsection{Training and Loss Functions}
\label{sec:loss}

We build the baseline training with the architecture proposed by  \cite{li2017improving}.
We explore multiple loss functions.
Ranking is shown to be a good loss for highly unbalanced multi-labelling problems~\cite{Usunier:2009:ROW:1553374.1553509,37180}.
One choice would be the hinge ranking loss \cite{weston2011wsabie,gong2013deep}:  
\begin{equation}
L_{hinge} = \max_{v\notin Y, u \in Y} \left ( 0, 1 + f_v(x) - f_u(x) \right )~,
\label{eq:hinge_rank_loss}
\end{equation}

where $f(x) : \mathbb{R}^d \rightarrow \mathbb{R}^K$ is a label (attribute) prediction model that maps an image to a K-dimensional label space which represents the confidence scores. The loss function is designed such that $f(x)$ produces a vector whose values for true labels are greater than those for negative labels (i.e. $f_u(x)>f_v(x),$ $\forall u \in Y, v \notin Y$). This creates the framework of learning to rank~\cite{liu2009learning} via pairwise comparisons.

However, the hinge loss (Eq.~\ref{eq:hinge_rank_loss}) is shown~\cite{li2017improving} to be difficult to optimize since it is non-smooth and as a solution the authors in~\cite{li2017improving} proposed the smooth ranking loss \cite{li2017improving}:

\begin{equation}
 L_{log\_sum\_exp} = log\left( 1 + \sum_{v\notin Y, u \in Y} \exp \left (f_v(x) - f_u(x) \right ) \right)~,  
\label{eq:smooth_rank_loss}
\end{equation}

which is basically a smooth approximation of Eq.~\ref{eq:hinge_rank_loss} using the log\_sum\_exp pairwise function.

We present the results using the best performing smooth ranking loss.
We make experiments with different learning algorithms: SGD with momentum \cite{ruder2016overview} and Adam \cite{kingma2014adam}.

\subsection{Evaluation metrics}
\label{subsec:metric}
We use ranking-based average precision (AVGPREC) as our main evaluation metric \cite{furnkranz2008multilabel}.
This metric is designed to evaluate ranked retrieval results of labels.
For each image the algorithm returns a ranked list of labels. The average precision metric computes for each relevant label in the retrieval list the percentage of relevant labels that are ranked higher than itself, and it averages these percentages over all relevant labels. AVGPREC is defined as:

\begin{equation}
AVGPREC(P_{x},\tau )= \frac{1}{\left | P_{x} \right |}\sum_{\lambda\in P_{x}}\frac{\left | \left \{ {\lambda}'\in P_{x}|\tau({\lambda}')\leq \tau(\lambda) \right \} \right |}{\tau(\lambda)}~,
\label{eq:avgprec}
\end{equation}

where $P_{x}$ is the list of relevant labels for the given instance $x$. $\tau({\lambda}')$ and $\tau({\lambda})$ are the positions of the predicted ranking labels ${\lambda}'$ and $\lambda$, respectively, where  ${\lambda}'$ is ranked higher than $\lambda$ (therefore, its position is before $\lambda$). 

We finally average over all the images.

Additionally, we consider the average precision (AP) per individual attribute (tag).
To calculate this, we consider for all the images the value of the logit corresponding to such label. We then compute the precision and recall at different threshold levels, create a precision-recall curve, and then average the precision over different recall values.
However, we take into account the observation by~\cite{davis2006relationship,flach2015precision} which linear interpolation of the points on the precision-recall curve provides an overly-optimistic measure of classifier performance. Therefore, we adopt a variant implementation of AP \cite{schutze2008introduction,everingham2010pascal} which does not interpolate the precision-recall curve. We finally calculate the weighted mean  average precision (W\_MAP)~\cite{zhao2015deep}, weighting by the frequency of instances per label.

%-------------------------------------------------------------------------
\section{Data}
\label{sec:data}
The multi-label instance classification task requires a fine-grained dataset where each item is annotated with an ideally balanced number of individual labels, i.e. for each label a balanced and sufficient number of images should be provided. There are a few fine-grained datasets available~\cite{wah2011caltech,maji2013fine,krause20133d}. However, most of them do not include detailed part annotations. Since they are initially collected and prepared for fine-grained categorization, the only annotations provided are the classes that the images belong to; for instance, the images in  FGVC aircraft dataset~\cite{maji2013fine} are annotated with the model variant, family, and manufacturer names. 

On the other hand, datasets which are provided for multi-label classification at a high level of recognition such as COCO~\cite{lin2014microsoft} are not suitable since the items are not labeled at a fine-grained level.

CUB200 bird dataset~\cite{wah2011caltech} is a widely used dataset in fine-grained visual categorization domain. It incorporates 11788 images of 200 species of birds where each image is provided with several attributes. There are overall 312 attributes which make a vocabulary of the same size for our experiment. An example of this dataset along with its tags is shown in Figure~\ref{fig:teasor}.

In the original dataset, the id of each image is repeated per attribute (i.e. 312 separate lines are written for each image) and a binary value of $0$ or $1$ indicates the presence or absence of each attribute for the corresponding image. As usual with a multi-label image classification task we need to provide our network with labels for each training image. We generate a binary vector of attributes $[0, 1]$ of size 312 (total number of attributes) per image. Every index of vector holding the value of one represents the presence of that attribute and zero means the corresponding attribute is absent. These attributes/labels are extracted for each image in the training, test and validation set to be used as the input to the network. \footnote{The new format of CUB200bird train, test and validation sets are available (the link will be provided).} Here, we are not taking into account with what confidence the attribute is decided for each image. For the total 11788 images of the training set this provides us with a matrix of size $11788\times312$. We consider the same set of images as \cite{lin2015bilinear} for the training set with around 6000 images. From the total 5794 test images we separate around 700 images for validation set which leaves 5094 images for the test set.   

The 312 attributes in the CUB dataset \cite{wah2011caltech} are formed into 28 groups including bill\_shape, eye\_colour, wing\_shape, etc. Each group of attributes varies from three to 14 diversities. For instance, the bill\_shape could be in the form of curved, dagger, hooked, needle, hooked\_seabird, spatulate, all-purpose, cone or specialized. As it is the case with the underpart\_colour in Figure~\ref{fig:teasor} which is both blue and grey, some groups of attributes might have more than one diversity which makes the task of retrieving the attributes even harder and defines the problem as inherently multi-labelling.

%-------------------------------------------------------------------------
\section{Experiments}
\label{sec:exp}
We adopt the VGG16 architecture as a backbone for our multi-labelling experiments. We compare our FineTag architecture with VGG16 based architecture both using ranking loss to train (denoted as VGG16+rank in Figure~\ref{fig:net}).

Following what is commonly done in this field, the pre-trained weights on Imagenet are used for the classification task \cite{deng2009imagenet,krizhevsky2012imagenet} by VGG16. We repeated the experiments with two different optimizers: stochastic gradient descent with momentum and Adam's optimizer~\cite{kingma2014adam}. 

Similarly, we trained our Finetag architecture initializing the convolutional layers with the weights of VGG16 trained on ImageNet classification. We initialized the 1x1 convolutional filter (ICA in Figure~\ref{fig:net}) with coefficients obtained with a dimensionality reduction based ICA or alternatively on PCA.
We explored different number of components, from three to 100, and found 20 components to be optimal. Like with VGG16, we repeated the experiment with two optimizers: stochastic gradient descent with momentum and adam, the results for which are shown in rows two and five of Table~\ref{tab:ap} respectively.

Ultimately, to make a fairer comparison of the transfer learning exploited by our architecture, we repeated the experiment for VGG16+rank by using the pre-trained weights on Imagenet only for the convolutional layers of VGG16. We randomly initialized the remaining FC layers with Xavier initialization~\cite{glorot2010understanding}. This way we use transfer learning for the two architectures with the same amount of information.

The code is built on Tensorflow framework and the experiments are run on an NVIDIA Tesla V100 GPU. We train both networks for a few epochs. We found the batch size of 16 and learning rate of 0.00001 when using Adam optimizer and 0.0001 when using stochastic gradient descent with momentum to be the optimum settings for both networks. %The training (on the mentioned GPU) takes roughly 60 seconds and 90 seconds per epoch for FineTag and VGG16+rank, respectively. 

\subsection{Results}
Table~\ref{tab:ap} summarizes the results of weighted mean average precision (W\_MAP) over all labels as well as the ranking-based average precision (AVGPREC) over all images (see Section~\ref{subsec:metric}). Using both method of optimization FineTag architecture performs better. Moreover, FineTag is quite robust towards the choice of optimizer. We can see that even with stochastic gradient descent with momentum it is possible to get almost equal results as with Adam's, while the deep net's acceptable performance is very dependant on the type of optimizer. Further, a fairer comparison between FineTag architecture (last row of the table) and VGG16+rank architecture where the pre-trained weights on Imagenet are used only for initialization in the convolutional layers (second row from bottom) shows that FineTag architecture has a  significantly better performance on the CUB200 birds dataset.

\begin{table}[!h]
\centering
\caption{Weighted Mean Average Precision (W\_MAP) over all labels, and ranking-based average precision (AVGPREC) over all images.}
\label{tab:ap}
\begin{tabular}{l|l|l|l|l}
\hline
method           &  Imagenet initialisation     & optimizer & W\_MAP   & AVGPREC \\ \hline
VGG16+rank            & all layers  & momentum  & 0.44 &  0.48  \\ 
\textbf{FineTag} & conv layers		 & momentum  & \textbf{0.46} &  \textbf{0.59}  \\\hline
VGG16+rank            & all layers & Adam      & 0.50 &  0.58  \\ 
VGG16+rank            & conv layers & Adam      & 0.29 &   0.43 \\
\textbf{FineTag} &	conv layers				 & Adam      & \textbf{0.52} &  \textbf{0.60}  \\ \hline
\end{tabular}
\label{summary}
\end{table}

\begin{figure}[!h]
    \centering
    %trim={<left> <lower> <right> <upper>}
    \includegraphics[trim={2cm 0cm 1cm 0cm},clip, width=13cm]{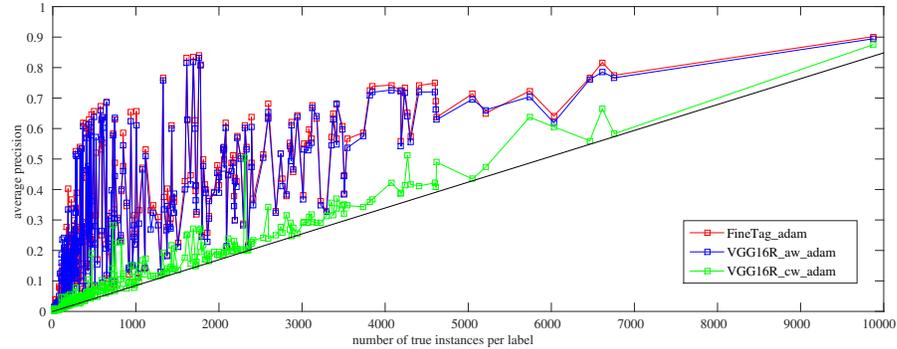}
    \caption{The relationship between the number of instances per label and the average precision. When there are few instances available for a label the average precision is very low for that label. On the other hand, with enough number of instances the label is guaranteed to be retrieved with an acceptable average precision.}
    \label{fig:APpersampledistribution}
\end{figure}

Table~\ref{tab:perlabel} shows the weighted mean average precision (W\_MAP) per group of labels. As mentioned before, there are 28 groups of attributes which are listed in the header of Table~\ref{tab:perlabel} (represented in two sections to fit the space). Each group of attributes could vary between three to 14 different diversities. The results in Table~\ref{tab:perlabel} are the  average over all diversities per group of attributes. To have a fair evaluation we have taken the number of images per label into account. A label could have as few as 15 images in the whole dataset and as many as almost 10000 images. Obviously, this is going to affect the training process and as it is often the case with imbalance data when the number of images is very few for a label, the mean average precision for that label is significantly lower (this is shown in Figure~\ref{fig:APpersampledistribution}). This is true for both networks trained with different optimizes. To compensate for the unbalanced distribution of instances across the labels we are multiplying the mean average precision for each label by a weight ~\cite{zhao2015deep} which is calculated as the frequency of instances for the corresponding label over the whole dataset.

\begin{table}[!h]
\centering
\caption{ Weighted mean  average precision (W\_MAP) per label is shown for per group of attributes. Each group of attribute in practice has between three to 14 varieties.}
\label{tab:perlabel}
\renewcommand{\arraystretch}{1.5}
\small
\resizebox{\textwidth}{!}{%
\begin{tabular}
{c|c| c| c| c| c| c| c| c| c| c| c| c| c| c}
\hline
method       & {\rotatebox[origin=c]{90}{bill\_shape}}   & {\rotatebox[origin=c]{90}{wing\_colour}}   & {\rotatebox[origin=c]{90}{upperpart\_colour}} & {\rotatebox[origin=c]{90}{underparts\_colour}} & {\rotatebox[origin=c]{90}{breast\_pattern}} & {\rotatebox[origin=c]{90}{back\_colour}}  & {\rotatebox[origin=c]{90}{tail\_shape}}   & {\rotatebox[origin=c]{90}{uppertail\_colour}} & {\rotatebox[origin=c]{90}{head\_pattern}} & {\rotatebox[origin=c]{90}{breast\_colour}} & {\rotatebox[origin=c]{90}{throat\_colour}} & {\rotatebox[origin=c]{90}{eye\_colour}}   & {\rotatebox[origin=c]{90}{bill\_length}}  & {\rotatebox[origin=c]{90}{forhead\_colour}}   \\ \hline
%----------------------------------------------------------------------------------------------------------------%
vgg16R\_aw\_mom  & 0.37          & 0.49          & 0.48              & 0.50               & 0.13            & 0.44          & 0.22          & 0.35              & 0.23          & 0.50           & 0.49           & 0.73          & 0.60        & 0.47\\%\hline %----------------------------------------------------------------------------------------------------------------%
\textbf{FineTag\_mom}   & 0.44          & 0.50          & 0.49              & 0.54               & 0.14            & 0.44          & 0.22      & 0.35              & 0.25          & 0.53           & 0.53           & 0.74          & 0.66      & 0.50\\ \hline
%----------------------------------------------------------------------------------------------------------------%
vgg16R\_aw\_adam & 0.52          & 0.55          & 0.54              & 0.56               & 0.14            & 0.49          & \textbf{0.25} & 0.38              & 0.29          & 0.56           & 0.56     &    \textbf{0.76} & 0.71          & 0.54   \\ %\hline
%----------------------------------------------------------------------------------------------------------------%
vgg16R\_cw\_adam & 0.33          & 0.29          & 0.27              & 0.25               & 0.10            & 0.22          & 0.22          & 0.19              & 0.17          & 0.24           & 0.24           & 0.72          & 0.57          & 0.22\\ %\hline
%----------------------------------------------------------------------------------------------------------------%
\textbf{FineTag\_adam}  & \textbf{0.54} & \textbf{0.57} & \textbf{0.55}     & \textbf{0.59}      & \textbf{0.15}   & \textbf{0.50} & 0.24          & \textbf{0.40}     & \textbf{0.30} & \textbf{0.58}  & \textbf{0.57}  & \textbf{0.76} & \textbf{0.73} & \textbf{0.56}       \\ \hline \hline
%----------------------------------------------------------------------------------------------------------------%
&{\rotatebox[origin=c]{90}{tail\_colour}}  & {\rotatebox[origin=c]{90}{nape\_colour}}  & {\rotatebox[origin=c]{90}{belly\_colour}} & {\rotatebox[origin=c]{90}{wing\_shape}}   & {\rotatebox[origin=c]{90}{size}}          & {\rotatebox[origin=c]{90}{shape}}         & {\rotatebox[origin=c]{90}{back\_pattern}} & {\rotatebox[origin=c]{90}{tail\_pattern}} & {\rotatebox[origin=c]{90}{belly\_pattern}} & {\rotatebox[origin=c]{90}{primary\_colour}} & {\rotatebox[origin=c]{90}{leg\_colour}}   & {\rotatebox[origin=c]{90}{bill\_colour}}  & {\rotatebox[origin=c]{90}{crown\_colour}} & {\rotatebox[origin=c]{90}{wing\_pattern}} \\ \hline
%----------------------------------------------------------------------------------------------------------------%
vgg16R\_aw\_mom  &     0.37          & 0.48          & 0.49          & 0.22          & 0.49          & 0.41          & 0.38          & 0.36          & 0.53           & 0.56            & 0.26          & 0.36          & 0.47          & 0.38\\
%----------------------------------------------------------------------------------------------------------------%
\textbf{FineTag\_mom}& 0.37          & 0.49          & 0.53          & 0.22          & 0.49          & 0.41          & 0.39          & 0.36          & 0.56           & 0.56            & 0.28          & 0.36          & 0.50          & 0.40\\ \hline
%----------------------------------------------------------------------------------------------------------------%
vgg16R\_aw\_adam             & 0.41          & 0.54          & 0.56          & \textbf{0.24} & 0.53          & 0.46          & 0.43          & 0.40          & 0.59           & \textbf{0.62}   & 0.31          & \textbf{0.42} & 0.55          & 0.46    \\     
%----------------------------------------------------------------------------------------------------------------%
vgg16R\_cw\_adam             & 0.22          & 0.23          & 0.24          & 0.22          & 0.49          & 0.37          & 0.28          & 0.28          & 0.43           & 0.28            & 0.20          & 0.28          & 0.22          & 0.30 \\
%----------------------------------------------------------------------------------------------------------------%
\textbf{FineTag\_adam}& \textbf{0.42} & \textbf{0.55} & \textbf{0.58} & \textbf{0.24} & \textbf{0.55} & \textbf{0.47} & \textbf{0.44} & \textbf{0.41} & \textbf{0.60}  & \textbf{0.62}   & \textbf{0.32} & 0.41          & \textbf{0.56} & \textbf{0.48}\\ \hline

\end{tabular}
}
\end{table}

In Table~\ref{tab:perlabel}, first row in each section holds the result for VGG16+rank architecture using Imagenet weights for initialization in all layers with momentum optimizer (denoted as vgg16R\_aw\_mom). In the second row in each section the results of FineTag architecture with momentum optimizer are shown (denoted as FineTag\_mom). The next two rows represent the results for VGG16+rank architecture with Adam optimizer once with initialising all layers with pre-trained Imagenet weights and once initializing only the convolutional layers (shown as vgg16R\_aw\_adam and vgg16R\_cw\_adam, respectively). Finally, the last row of the table holds the results for FineTag architecture with Adam optimizer (denoted as FineTag\_adam). We can see that FineTag returns a higher accuracy for most groups of attributes. Again, when comparing both networks trained with similar conditions, i.e. using the pre-trained weights on Imagenet only for convolutional layers, FineTag outperforms the VGG16+R architecture. (To make the analysis of this table visually simpler, we have only highlighted the best results regardless of the chosen optimizer).

As mentioned before, the FineTag architecture is a much smaller network in terms of number of parameters (almost $40\times$ smaller than VGG16+rank).

%-------------------------------------------------------------------------
\section{Summary}
\label{sec:sum}
In this paper, we introduced the FineTag architecture: a network for recognizing multiple tags of a single item (multi-label item classification task) at fine-grained level of details, a task which has been less addressed. We compared our proposed architecture with a deep net (VGG16) using a rank-based loss for classification. Our experiments on CUB200 bird dataset showed that FineTag outperforms the baseline architecture in almost every group of attributes. The architecture is fully convolutional, with a relatively high resolution feature maps. This allows to run it on arbitrary size images. Further, we showed that FineTag architecture has very few parameters compared to the baseline architecture ($40\times$ less). This is a highly important characteristic for the architecture to be deployed in a retrieval system. Moreover, being a shallower network, FineTag is not very dependant on the type of optimizer and can perform well enough with a simple optimizer like stochastic gradient descent with momentum.

One challenge which is remained with multi-label item classification task at fine-grained level is the lack of suitable datasets. Training a network for fine tagging requires a well-balanced densely labelled dataset. For our experiments, we adopted CUB200 birds dataset~\cite{wah2011caltech} to a format which is suitable for the network. However, even with CUB200 bird dataset the number of images per label is not well proportioned and it varies from 15 to 10000 images. In a more accurate experiment, we would like to address this issue by creating better datasets. Labelling the CUB200 bird dataset was a big challenge since it requires expert opinions for labelling, but other datasets such as the aircraft dataset~\cite{maji2013fine} could be labelled accurately for future experiments without expert knowledge.

\bibliographystyle{splncs}
\bibliography{egbib}

%this would normally be the end of your paper, but you may also have an appendix
%within the given limit of number of pages
\end{document}